\documentclass{article} %
\usepackage{nips14submit_e,times}
\usepackage{url}
\usepackage{graphicx}
\usepackage[font={small}]{caption}
\usepackage{subcaption}
\usepackage{amsmath,amssymb}
\usepackage{floatrow}
\usepackage{array}
\usepackage{booktabs}
\usepackage{makecell}
\usepackage{setspace}

\newfloatcommand{capbtabbox}{table}[][\FBwidth]

\title{Towards a Visual Turing Challenge}%

\author{
Mateusz Malinowski \hspace{50pt} Mario Fritz
\\
Max Planck Institute for Informatics \\
Saarbr{\"u}cken, Germany \\
\texttt{\{mmalinow,mfritz\}@mpi-inf.mpg.de} 
}

\nipsfinalcopy %

\begin{document}

\maketitle

\begin{abstract}
As language and visual understanding by machines progresses rapidly,  we are observing an increasing
interest in holistic architectures that tightly interlink both modalities in a joint learning and inference process.
This trend has allowed the community to progress towards more challenging and open tasks and refueled the hope at achieving the old AI dream of building machines that could pass a turing test in open domains.
In order to steadily make progress towards this goal, we realize that quantifying performance becomes increasingly difficult. Therefore we ask how we can precisely define such challenges and
how we can evaluate different algorithms on this open tasks? 
In this paper, we summarize and discuss such challenges as well as try to give answers where appropriate options are available in the literature.
We exemplify some of the solutions on a recently presented dataset of  question-answering task based on real-world indoor images that establishes a visual turing challenge.
Finally, we argue despite the success of unique ground-truth annotation, we likely have to step away from carefully curated dataset and rather rely on 'social consensus' as the main driving force to create suitable benchmarks. Providing coverage in this inherently ambiguous output space is an emerging challenge that we face in order to make quantifiable progress in this area.
\end{abstract}
\section{Introduction}
\label{section:introduction}
Recently we witness a tremendous progress in the machine perception \cite{krizhevsky2012imagenet, gupta2014learning, girshick2014rcnn, pishchulin2013strong,tompson2014joint, he2014spatial,lee2014deeply,simonyan2014very} and in the language understanding \cite{BlackburnBos:2005,zettlemoyer2007online, kwiatkowski2010inducing, mikolov2013distributed,cho2014learning} tasks.
The progress in both fields has inspired researchers to build holistic architectures for challenging grounding  \cite{matuszek2012joint, krishnamurthy2013jointly}, natural language generation from image/video \cite{farhadi2010every,kulkarni2011baby,senina2014coherent}, image-to-sentence alignment \cite{socher2013grounded,karpathy2014deep,mao2014explain,kong2014you}, and recently presented question-answering problems \cite{liang2013learning,berant2014semantic,iyyer2014neural,faderopen,malinowski14nips}. 
In this paper we argue for a Visual Turing Test - an open domain task of question-answering based on real-world images that resemblances the famous Turing Test \cite{turing1950computing,lacurts2011criticisms} and deviates from  other attempts \cite{shan2013visual,lake2013one,battaglia2013simulation} - and discuss challenges together with tools to benchmark different models on such task. 

We typically measure the progress in the field by quantifying the performance of different methods against a carefully crafted set of benchmarks. 
Crowdsourcing in combination of machine learning approaches have served us well to generate curated datasets with a unique ground truth at scale \cite{welinder2010cvprw,welinder2010nips}.
As the complexity and the openness of the task grows, the quest of crafting good benchmarks also becomes more difficult. First, interpreting and evaluating the answer of a system becomes increasingly difficult and ideally would rely on human judgement. Yet we want to have objective metrics that we can evaluate automatically at large scale. Second, establishing an evaluation methodology that assigns scores over a large output domain is challenging, as any system based on ontologies will have limited coverage. Third, if our aim is to mimic human response, we have to deal with inherent ambiguities due to human judgement that stem from issues like binding, reference frames, social conventions.
For instance \cite{malinowski14nips} reports that for a question answering task on real-world images even human answers are inconsistent. Obviously this cannot be a problem of humans but rather argues for inherent ambiguities in the task.

Competing methods are validated against true annotations, but what is the ``truth'' in a task where even human answers cannot completely agree with each other? Instead of seeking an unique, ``true'' answer we suggest to look into 'social consensus' that takes multiple human answers as different interpretations of the question into account. This enables us to incorporate 'agreement' between the humans directly into the metric. Although the idea is not entirely new \cite{amfm_pami2011,hodosh2013framing,farhadi2010eccv}, we believe it sits at the core of building more open and holistic challenges. 

We exemplify some of our findings on the DAQUAR dataset \cite{malinowski14nips} with the aim of demonstrating different challenges that are present in the dataset. 
We hope that our exposition is helpful towards building a public visual turing challenge and will generate a discussion for the agreeable evaluation procedure and designing systems that can address open domain tasks.

In this paper holistic architecture (also holistic learner) is a machine learning architecture designed to work on the task that fuses at least two modalities, e.g. language and vision. 
The external world is a part of a task accessible to the holistic learner only via sensors and it can be either human world (the world that surrounds us), or a machine world that models some aspects of human world.

\section{Challenges}
As we strive for more holistic and open tasks such as grounding or question-answering based on images, we need to deal with a large gamut of challenges.
In this section we have distilled and discuss some of the most prominent ones in order to guide the further discussion.

\paragraph{Vision and language}
{\it Scalability:} Perception and natural language understanding are crucial parts of holistic reasoning as they ground any representation in the external world and therefore serve as a common reference point for machines and humans. The human conceptualization divides these percepts into different instances, categories as well as spatio-temporal concepts. Architectures that aim at mimicking or reproducing this space of human concepts need to capture the same diversity and therefore scale up to thousands of concepts \cite{wsabie,perronnin2012towards,Hoffman14Lsda}. \\
{\it Concept ambiguity:} As the number of categories grows, the semantic boundaries become more fuzzy, and hence ambiguities are inherently introduced \cite{lakoff1990women,deng2010does}. For instance, sometimes we may overlook the difference between 'night stand' and 'cabinet', or 'armchair' and 'sofa'. Therefore it is reasonable to expect from the holistic architectures to create alternative hypotheses of the external world during inference. 
This also relates to the gradual category membership in human perception as portrayed in the prototype theory \cite{lakoff1990women,rosch1973natural}.
\\
{\it Attributes: }The human concepts are not  limited to  object categories, but also include attributes such as genders, colors, states (lights can be either on or off).
Often these concepts cannot be learned on their own, but rather are contextualized by the associated noun. E.g. white in ``white'' elephant is surly different from ``white'' in white snow.\\
{\it Ambiguity in reference resolution:} 
Reliably answering on questions is challenging even for humans. The quality of an answer depends on how ambiguous and latent notions of reference frames and intentions are understood \cite{malinowski14nips,golland2010game}.
Depending on the cultural bias and the context, we may use object-centric or observer-centric or even world-centric frames of reference \cite{levinson2003space}. Moreover, it is even unclear what 'with', 'beneath', 'over' mean. It seems at least difficult to symbolically define them in terms of predicates. 
While holistic learning and inference encompassing all the aforementioned aspects has yet to be shown, current research directions show promise \cite{beltagy2013montague,rocktaschellow,lewiscombining} by adapting the symbolic-based approaches  \cite{zettlemoyer2007online,kwiatkowski2010inducing,liang2013learning,berant2014semantic} with vector-based approaches \cite{mikolov2013distributed,socher2013grounded, iyyer2014neural} to represent the meaning. 

\paragraph{Common sense knowledge}
It turns out that some questions can solely be answered with the access to common sense knowledge with high reliability. For instance "Which object on the table is used for cutting?" already narrows the likely options significantly and the correct answer is probably ``knife'' or ``scissors''. Other questions like "Which hand of the teacher is on her chin?" require the mixture of the vision and language. To understand the question, a holistic learner needs to first detect a person, figure out that the person may be a teacher, understand a gender of the person, detect her chin, understand 'left' and 'right' side, and finally relates 'her' with the 'teacher'. 

However, different parts of the common sense knowledge can be used with different modality.
An 'object for cutting' is not about seeing but about the affordance of the object and it cannot be learnt solely from the set of images. On the other hand things that often co-occur together may stand for the visual-based common sense knowledge. For instance we may expect to find a scissor or a pen inside a small plastic box, but never a wall or a window. 

Common sense knowledge can help holistic machine learning architectures to either fulfill the task (question "Which object on the table is used for cutting?" can utilizes this type of knowledge), or limit the hypothesis space and hence to reduce the computational complexity of the search problem. For instance an architecture could be guided by its common sense knowledge to limit the space of possible locations of the 'scissors' and answer on  "What is in front of scissors?" more effectively.

\paragraph{Defining a benchmark dataset and quantifying performance}
We argue that the question answering based on the visual input task significantly differ from the grounding problem and has unique advantages towards defining a challenge dataset. Most prominently, the latter is about finding (either with a hand-crafted set of rules or learnt-based approaches) a mapping between the linguistic fragments and the physical world \cite{matuszek2012joint,krishnamurthy2013jointly, harnad1990symbol}, whereas the question answering task is about an end-to-end system where we do not necessarily want to enforce any constraints or penalty for the internal representation of the holistic learner. In this sense grounding is a latent sub-task that the holistic learner needs to solve, but will not be evaluated on.
Finally, we argue that establishing benchmark dataset based on a question answering task similar to a turing test, is more tractable. Learning grounding asks for  exhaustive symbolic-based annotations of the world, while question answering only needs textual annotations for the aspects that the question refers to.
\section{DAQUAR: Building a Dataset for Visual Turing Challenge}
DAQUAR \cite{malinowski14nips} is a challenging, large dataset for a question answering task based on real-world images. The images present real-world indoor scenes \cite{silbermanECCV12}, while the questions are unconstrained natural language sentences. DAQUAR's language scope is beyond the nouns or tuples that are typical to  recognition datasets \cite{ILSVRCarxiv14,rohrbach2011evaluating,LanYWM12}. Other, linguistically rich datasets either do not tackle images at all \cite{zelle1996learning,berant2013semantic} or consider only few in very constrained domain \cite{krishnamurthy2013jointly}, or are more suitable for the learning an embedding/image-sentence retrieval or language generation \cite{kong2014you,rashtchian2010collecting,rohrbach2012script,gong2014improving}. 
In this section we discuss in isolation different challenges reflected in DAQUAR. 
\paragraph{Vision and language}
The machine world in DAQUAR is represented as a set of images and questions about their content.
DAQUAR contains $1088$ different nouns in the question, $803$ in the answers, and $1586$ altogether (we use the Stanford POS Tagger \cite{toutanova2003feature} to extract the nouns from the questions).
If we consider only nouns in singular form in the questions, we still have $573$ categories.
The current state-of-the-art semantic segmentation methods on the NYU-Depth V2 dataset \cite{silbermanECCV12} can discriminate only between up to $37$ object categories \cite{gupta2014learning,lin2013holistic,gupta2013perceptual}, much fewer to what is needed. DAQUAR also contains other parts of speech where only colors and spatial prepositions are grounded in \cite{malinowski14nips}. 

Moreover, ambiguities naturally emerge due to fine grained categories that exist in DAQUAR.
For instance 'night stand', 'stool' and 'cabinet' sometimes refer to the same thing. There is also a variation in the naming of colors among the annotations. 
Questions rely heavily on the spatial concepts with different frame of reference.

DAQUAR includes various challenges related to natural language understanding.
Any semantic representation needs to work with the large number of predicates (reaching about $4$ million to account different interpretations of the external world), with questions of substantial length ($10.5$ words in average with variance $5.5$; the longest question has $30$ words), and possible language errors in the questions.

\paragraph{Common sense knowledge}
DAQUAR includes questions that can be reliably answered using common sense knowledge. For instance "Which object on the table is used for cutting?" already provides strong non-visual cues for the ``cutting'' object.
Answers on other questions, such as "What is above the desk in front of scissors?", can be improved if the search space is reasonable restricted.
Moreover, some annotators hypothesize missing parts of the object based on their common sense. To sum up, we believe that common sense knowledge is an interesting venue to explore with DAQUAR.

\paragraph{Question answering task}
The question answering task is also about understanding hidden intentions of the questioner with grounding as a sub-goal to solve. Some authors \cite{liang2013learning, berant2014semantic, malinowski14nips} treat the grounding (understood here as the logical representation of the meaning of the question) as a latent variable in the question answering task.
Others \cite{golland2010game} have modeled the pragmatic effects in the question answering task, but such approaches have never been shown to work in less constrained environments. 
\section{Quantifying the Performance of Holistic Architectures}
\label{section:benchmarks}
Together with increasing complexity and openness of the task, quantifying performance of the holistic architectures becomes challenging due to several issues:
\\
{\it Automation:}
Evaluating answers on such complex tasks as answering on questions requires a quite deep understanding of natural language, involved concepts and hidden intentions of the questioner. The ideal but impractical metric would be to manually judge every single answer of every architecture individually. Since this is infeasible we are seeking an automatic approximation so that we can evaluate different holistic architectures at scale.
\\
{\it Ambiguity:}
The complex tasks that we are interested in are inherently ambiguous. The ambiguities stem from cultural bias, different frame of reference and fined grained categorization. This implies that multiple interpretations of a question are possible and hence many correct answers. 
\\
{\it Coverage:} 
Since there are multiple ways of expressing the same concept, the automatic performance metric should take the equivalence class among the answers into the consideration by assigning similar scores to all members of the same class. There are attempts to alleviate this issue via defining similarity scores \cite{wu1994verbs} over the lexical databases \cite{miller1995wordnet, fellbaum1999wordnet}. These approaches, however, lacks of coverage: we cannot assign a similarity between the terms that are not represented in the structure.

\paragraph{WUPS scores}
We exemplify the aforementioned requirements by illustrating the WUPS score - an automatic metric that quantifies performance of the holistic architectures proposed by \cite{malinowski14nips}. This metric is motivated by the development of a 'soft' generalization of accuracy that takes ambiguities of different concepts into account via the set membership measure $\mu$:
\begin{align}
\label{eq:wups_score}
\frac{1}{N} \sum_{i=1}^N \min\{ \prod_{a \in A^i} \max_{t\in T^i} \mu(a, t) ,\; \prod_{t \in T^i} \max_{a \in A^i} \mu(a, t)\} \cdot 100
\end{align}
where for each $i$-th question, $A^i$ and $T^i$ are the answers produced by the architecture and human respectively, and they are represented as bags of words. 
The authors of \cite{malinowski14nips} have proposed using WUP similarity \cite{wu1994verbs} as the membership measure $\mu$ in the WUPS score. Such choice of $\mu$ suffers from the aforementioned coverage problem and the whole metric takes only one human interpretation of the question into account.

\paragraph{Future directions for defining metrics}
Recent work provides several directions towards improving scores.
To deal with  ambiguities that stem from different readings of the same question we are collecting more human answers per question and we propose, based on that, two generalizations of WUPS score. The first, we call Interpretation Metric, runs Eq. \ref{eq:wups_score} over many human answers and takes the maximal score, so that
the machine answer is high if it is similar to at least one human answer. However, with many human answers, we can also rank higher the machine answers that are  'socially agreeable' by measuring if they agree with most human answers. This can be done by averaging over multiple human answers. We call such second extension, Consensus Metric. 
The problem with coverage can be potentially alleviated with vector based representations \cite{mikolov2013distributed} of the answers. Although in this case the coverage issues are less problematic, we understand the concerns that such score is dependent on the training data used to build such representation. On the other hand, due to abundance of textual data and recent improvements of vector based approaches \cite{mikolov2013distributed,pennington2014glove}, we consider it as a valid alternative to similarities that are based on ontologies.

\paragraph{Experimental scenarios}
In many cases, success on challenging learning problems has been accelerated by use of external data in the training, e.g. in object detection \cite{girshick2014rcnn}.
We believe that a Visual Turing challenge should consists of a sub-task with a prohibited use of auxiliary data to understand how the holistic learners generalize from limited and challenging data in a more established setup. On the other hand we should not limit ourselves to such artificial restrictions in building next generation of the holistic learners. Therefore open sub-tasks with a permissible use of another sources in the training have to be stated, including: additional vision and language resources, synthetic data and curated questions. 
\section{Summary}
The goal of this contribution is to sparkle the discussions about benchmarking holistic architectures on complex and more open tasks. 
We identify particular challenges that holistic tasks should exhibit and exemplify how they are manifested in a recent question answering challenge \cite{malinowski14nips}. 
To judge competing architectures and measure the progress on the task, we suggest several directions to further improve existing metrics, and discuss different experimental scenarios. 
{\bf Acknowledgement:} We would like to thank Michael Stark for his comments on the draft.

\bibliographystyle{splncs}
\small
\bibliography{egbib}

\begin{thebibliography}{10}

\bibitem{krizhevsky2012imagenet}
Krizhevsky, A., Sutskever, I., Hinton, G.E.:
\newblock Imagenet classification with deep convolutional neural networks.
\newblock In: NIPS. (2012)

\bibitem{gupta2014learning}
Gupta, S., Girshick, R., Arbel{\'a}ez, P., Malik, J.:
\newblock Learning rich features from rgb-d images for object detection and
  segmentation.
\newblock In: ECCV.
\newblock (2014)

\bibitem{girshick2014rcnn}
Girshick, R., Donahue, J., Darrell, T., Malik, J.:
\newblock Rich feature hierarchies for accurate object detection and semantic
  segmentation.
\newblock In: CVPR. (2014)

\bibitem{pishchulin2013strong}
Pishchulin, L., Andriluka, M., Gehler, P., Schiele, B.:
\newblock Strong appearance and expressive spatial models for human pose
  estimation.
\newblock In: ICCV. (2013)

\bibitem{tompson2014joint}
Tompson, J., Jain, A., LeCun, Y., Bregler, C.:
\newblock Joint training of a convolutional network and a graphical model for
  human pose estimation.
\newblock In: NIPS. (2014)

\bibitem{he2014spatial}
He, K., Zhang, X., Ren, S., Sun, J.:
\newblock Spatial pyramid pooling in deep convolutional networks for visual
  recognition.
\newblock In: ECCV.
\newblock (2014)

\bibitem{lee2014deeply}
Lee, C.Y., Xie, S., Gallagher, P., Zhang, Z., Tu, Z.:
\newblock Deeply-supervised nets.
\newblock arXiv:1409.5185 (2014)

\bibitem{simonyan2014very}
Simonyan, K., Zisserman, A.:
\newblock Very deep convolutional networks for large-scale image recognition.
\newblock arXiv:1409.1556 (2014)

\bibitem{BlackburnBos:2005}
Blackburn, P., Bos, J.:
\newblock Representation and Inference for Natural Language. A First Course in
  Computational Semantics.
\newblock CSLI (2005)

\bibitem{zettlemoyer2007online}
Zettlemoyer, L.S., Collins, M.:
\newblock Online learning of relaxed ccg grammars for parsing to logical form.
\newblock In: EMNLP-CoNLL. (2007)

\bibitem{kwiatkowski2010inducing}
Kwiatkowski, T., Zettlemoyer, L., Goldwater, S., Steedman, M.:
\newblock Inducing probabilistic ccg grammars from logical form with
  higher-order unification.
\newblock In: EMNLP. (2010)

\bibitem{mikolov2013distributed}
Mikolov, T., Sutskever, I., Chen, K., Corrado, G.S., Dean, J.:
\newblock Distributed representations of words and phrases and their
  compositionality.
\newblock In: NIPS. (2013)

\bibitem{cho2014learning}
Cho, K., van Merrienboer, B., Gulcehre, C., Bougares, F., Schwenk, H.,
  Bahdanau, D., Bengio, Y.:
\newblock Learning phrase representations using rnn encoder-decoder for
  statistical machine translation.
\newblock In: EMNLP. (2014)

\bibitem{matuszek2012joint}
Matuszek, C., Fitzgerald, N., Zettlemoyer, L., Bo, L., Fox, D.:
\newblock A joint model of language and perception for grounded attribute
  learning.
\newblock In: ICML. (2012)

\bibitem{krishnamurthy2013jointly}
Krishnamurthy, J., Kollar, T.:
\newblock Jointly learning to parse and perceive: Connecting natural language
  to the physical world.
\newblock TACL (2013)

\bibitem{farhadi2010every}
Farhadi, A., Hejrati, M., Sadeghi, M.A., Young, P., Rashtchian, C.,
  Hockenmaier, J., Forsyth, D.:
\newblock Every picture tells a story: Generating sentences from images.
\newblock In: ECCV.
\newblock (2010)

\bibitem{kulkarni2011baby}
Kulkarni, G., Premraj, V., Dhar, S., Li, S., Choi, Y., Berg, A.C., Berg, T.L.:
\newblock Baby talk: Understanding and generating simple image descriptions.
\newblock In: CVPR. (2011)

\bibitem{senina2014coherent}
Rohrbach, A., Rohrbach, M., Qiu, W., Friedrich, A., Amin, S., Andriluka, M.,
  Pinkal, M., Schiele, B.:
\newblock Coherent multi-sentence video description with variable level of
  detail.
\newblock In: GCPR. (2014)

\bibitem{socher2013grounded}
Socher, R., Karpathy, A., Le, Q., Manning, C., Ng, A.:
\newblock Grounded compositional semantics for finding and describing images
  with sentences.
\newblock In: TACL. (2014)

\bibitem{karpathy2014deep}
Karpathy, A., Joulin, A., Fei-Fei, L.:
\newblock Deep fragment embeddings for bidirectional image sentence mapping.
\newblock In: NIPS. (2014)

\bibitem{mao2014explain}
Mao, J., Xu, W., Yang, Y., Wang, J., Yuille, A.L.:
\newblock Explain images with multimodal recurrent neural networks.
\newblock arXiv:1410.1090 (2014)

\bibitem{kong2014you}
Kong, C., Lin, D., Bansal, M., Urtasun, R., Fidler, S.:
\newblock What are you talking about? text-to-image coreference.
\newblock In: CVPR. (2014)

\bibitem{liang2013learning}
Liang, P., Jordan, M.I., Klein, D.:
\newblock Learning dependency-based compositional semantics.
\newblock Computational Linguistics (2013)

\bibitem{berant2014semantic}
Berant, J., Liang, P.:
\newblock Semantic parsing via paraphrasing.
\newblock In: ACL. (2014)

\bibitem{iyyer2014neural}
Iyyer, M., Boyd-Graber, J., Claudino, L., Socher, R., III, H.D.:
\newblock A neural network for factoid question answering over paragraphs.
\newblock In: EMNLP. (2014)

\bibitem{faderopen}
Fader, A., Zettlemoyer, L., Etzioni, O.:
\newblock Open question answering over curated and extracted knowledge bases.
\newblock In: KDD. (2014)

\bibitem{malinowski14nips}
Malinowski, M., Fritz, M.:
\newblock A multi-world approach to question answering about real-world scenes
  based on uncertain input.
\newblock In: NIPS. (2014)

\bibitem{turing1950computing}
Turing, A.M.:
\newblock Computing machinery and intelligence.
\newblock Mind (1950)  433--460

\bibitem{lacurts2011criticisms}
LaCurts, K.:
\newblock Criticisms of the turing test and why you should ignore (most of)
  them.
\newblock (2011)

\bibitem{shan2013visual}
Shan, Q., Adams, R., Curless, B., Furukawa, Y., Seitz, S.M.:
\newblock The visual turing test for scene reconstruction.
\newblock In: 3DV. (2013)

\bibitem{lake2013one}
Lake, B.M., Salakhutdinov, R., Tenenbaum, J.:
\newblock One-shot learning by inverting a compositional causal process.
\newblock In: NIPS. (2013)

\bibitem{battaglia2013simulation}
Battaglia, P.W., Hamrick, J.B., Tenenbaum, J.B.:
\newblock Simulation as an engine of physical scene understanding.
\newblock Proceedings of the National Academy of Sciences \textbf{110}(45)
  (2013)  18327--18332

\bibitem{welinder2010cvprw}
Welinder, P., Perona, P.:
\newblock Online crowdsourcing: rating annotators and obtaining cost-effective
  labels.
\newblock In: CVPR Workshops. (2010)

\bibitem{welinder2010nips}
Welinder, P., Branson, S., Perona, P., Belongie, S.J.:
\newblock The multidimensional wisdom of crowds.
\newblock In: NIPS. (2010)

\bibitem{amfm_pami2011}
Arbelaez, P., Maire, M., Fowlkes, C., Malik, J.:
\newblock Contour detection and hierarchical image segmentation.
\newblock TPAMI \textbf{33}(5) (May 2011)  898--916

\bibitem{hodosh2013framing}
Hodosh, M., Young, P., Hockenmaier, J.:
\newblock Framing image description as a ranking task: Data, models and
  evaluation metrics.
\newblock JAIR \textbf{47} (2013)  853--899

\bibitem{farhadi2010eccv}
Farhadi, A., Hejrati, M., Sadeghi, M.A., Young, P., Rashtchian, C.,
  Hockenmaier, J., Forsyth, D.:
\newblock Every picture tells a story: Generating sentences from images.
\newblock In: ECCV. (2010)

\bibitem{wsabie}
Weston, J., Bengio, S., Usunier, N.:
\newblock Wsabie: Scaling up to large vocabulary image annotation.
\newblock In: IJCAI. (2011)

\bibitem{perronnin2012towards}
Perronnin, F., Akata, Z., Harchaoui, Z., Schmid, C.:
\newblock Towards good practice in large-scale learning for image
  classification.
\newblock In: CVPR. (2012)

\bibitem{Hoffman14Lsda}
Hoffman, J., Guadarrama, S., Tzeng, E., Hu, R., Donahue, J., Girshick, R.,
  Darrell, T., Saenko, K.:
\newblock {LSDA}: Large scale detection through adaptation.
\newblock In: NIPS. (2014)

\bibitem{lakoff1990women}
Lakoff, G.:
\newblock Women, fire, and dangerous things: What categories reveal about the
  mind.
\newblock Cambridge Univ Press (1990)

\bibitem{deng2010does}
Deng, J., Berg, A.C., Li, K., Fei-Fei, L.:
\newblock What does classifying more than 10,000 image categories tell us?
\newblock In: ECCV. (2010)

\bibitem{rosch1973natural}
Rosch, E.H.:
\newblock Natural categories.
\newblock Cognitive psychology \textbf{4}(3) (1973)  328--350

\bibitem{golland2010game}
Golland, D., Liang, P., Klein, D.:
\newblock A game-theoretic approach to generating spatial descriptions.
\newblock In: EMNLP. (2010)

\bibitem{levinson2003space}
Levinson, S.C.:
\newblock Space in language and cognition: Explorations in cognitive diversity.
  Volume~5.
\newblock Cambridge University Press (2003)

\bibitem{beltagy2013montague}
Beltagy, I., Chau, C., Boleda, G., Garrette, D., Erk, K., Mooney, R.:
\newblock Montague meets markov: Deep semantics with probabilistic logical
  form.
\newblock In: *SEM. (2013)

\bibitem{rocktaschellow}
Rockt{\"a}schel, T., Bosnjak, M., Singh, S., Riedel, S.:
\newblock Low-dimensional embeddings of logic.
\newblock In: ACL Workshop on Semantic Parsing. (2014)

\bibitem{lewiscombining}
Lewis, M., Steedman, M.:
\newblock Combining formal and distributional models of temporal and
  intensional semantics.
\newblock In: ACL Workshop on Semantic Parsing. (2014)

\bibitem{harnad1990symbol}
Harnad, S.:
\newblock The symbol grounding problem.
\newblock Physica D: Nonlinear Phenomena \textbf{42}(1) (1990)  335--346

\bibitem{silbermanECCV12}
Silberman, N., Hoiem, D., Kohli, P., Fergus, R.:
\newblock Indoor segmentation and support inference from rgbd images.
\newblock In: ECCV. (2012)

\bibitem{ILSVRCarxiv14}
Russakovsky, O., Deng, J., Su, H., Krause, J., Satheesh, S., Ma, S., Huang, Z.,
  Karpathy, A., Khosla, A., Bernstein, M., Berg, A.C., Fei-Fei, L.:
\newblock Imagenet large scale visual recognition challenge.
\newblock arXiv:1409.0575 (2014)

\bibitem{rohrbach2011evaluating}
Rohrbach, M., Stark, M., Schiele, B.:
\newblock Evaluating knowledge transfer and zero-shot learning in a large-scale
  setting.
\newblock In: CVPR. (2011)

\bibitem{LanYWM12}
Lan, T., Yang, W., Wang, Y., Mori, G.:
\newblock Image retrieval with structured object queries using latent ranking
  svm.
\newblock In: ECCV. (2012)

\bibitem{zelle1996learning}
Zelle, J.M., Mooney, R.J.:
\newblock Learning to parse database queries using inductive logic programming.
\newblock In: AAAI. (1996)

\bibitem{berant2013semantic}
Berant, J., Chou, A., Frostig, R., Liang, P.:
\newblock Semantic parsing on freebase from question-answer pairs.
\newblock In: EMNLP. (2013)

\bibitem{rashtchian2010collecting}
Rashtchian, C., Young, P., Hodosh, M., Hockenmaier, J.:
\newblock Collecting image annotations using amazon's mechanical turk.
\newblock In: NAACL HLT Workshop. (2010)

\bibitem{rohrbach2012script}
Rohrbach, M., Regneri, M., Andriluka, M., Amin, S., Pinkal, M., Schiele, B.:
\newblock Script data for attribute-based recognition of composite activities.
\newblock In: ECCV.
\newblock (2012)

\bibitem{gong2014improving}
Gong, Y., Wang, L., Hodosh, M., Hockenmaier, J., Lazebnik, S.:
\newblock Improving image-sentence embeddings using large weakly annotated
  photo collections.
\newblock In: ECCV.
\newblock (2014)

\bibitem{toutanova2003feature}
Toutanova, K., Klein, D., Manning, C.D., Singer, Y.:
\newblock Feature-rich part-of-speech tagging with a cyclic dependency network.
\newblock In: HLT-NAACL. (2003)

\bibitem{lin2013holistic}
Lin, D., Fidler, S., Urtasun, R.:
\newblock Holistic scene understanding for 3d object detection with rgbd
  cameras.
\newblock In: ICCV. (2013)

\bibitem{gupta2013perceptual}
Gupta, S., Arbelaez, P., Malik, J.:
\newblock Perceptual organization and recognition of indoor scenes from rgb-d
  images.
\newblock In: CVPR. (2013)

\bibitem{wu1994verbs}
Wu, Z., Palmer, M.:
\newblock Verbs semantics and lexical selection.
\newblock In: ACL. (1994)

\bibitem{miller1995wordnet}
Miller, G.A.:
\newblock Wordnet: a lexical database for english.
\newblock CACM (1995)

\bibitem{fellbaum1999wordnet}
Fellbaum, C.:
\newblock WordNet.
\newblock Wiley Online Library (1999)

\bibitem{pennington2014glove}
Pennington, J., Socher, R., Manning, C.D.:
\newblock Glove: Global vectors for word representation.
\newblock In: EMNLP. (2014)

\end{thebibliography}

\end{document}